\documentclass{esannV2}
\usepackage[dvips]{graphicx}
\usepackage[latin1]{inputenc}
\usepackage{amssymb,amsmath,array}
\usepackage{url}            
\usepackage{booktabs}       
\usepackage{amsfonts}       
\usepackage{nicefrac}       
\usepackage[shortlabels]{enumitem}
\setlist[enumerate]{noitemsep}
\usepackage{color}
\usepackage{graphicx}
\usepackage{tabularx}
\usepackage{caption}
\usepackage{subcaption}
\usepackage{ragged2e}
\usepackage{multirow}

\begin{document}
\title{Structure Optimization for Deep Multimodal Fusion Networks using Graph-Induced Kernels}

\author{Dhanesh Ramachandram\thanks{These authors contributed equally} $^1$, Michal Lisicki$^{*1}$, Timothy J. Shields$^{*2}$, \\ Mohamed R. Amer$^2$ and Graham W.
Taylor$^1$
%
\vspace{.3cm}
%
\\1- Machine Learning Research Group, School of Engineering \\
University of Guelph, Canada.
%
\vspace{.1cm}
\\2- Center for Vision Technologies \\
SRI International, USA.\\
}

\maketitle

\begin{abstract}
A popular testbed for deep learning has been multimodal recognition of human activity or gesture involving diverse
inputs such as video, audio, skeletal pose and depth images. Deep learning architectures have excelled on such problems
due to their ability to combine modality representations at different levels of nonlinear feature extraction. However,
designing an optimal architecture in which to fuse such learned representations has largely been a non-trivial human
engineering effort. We treat fusion structure optimization as a hyper-parameter search and cast it as a discrete
optimization problem under the Bayesian optimization framework. We propose a novel graph-induced kernel to compute structural
similarities in the search space of tree-structured multimodal architectures and demonstrate its effectiveness using two
challenging multimodal human activity recognition datasets.
\end{abstract}

\section{Introduction}\label{sec:introduction}

With the increasing complexity of deep architectures
(e.g.~\cite{Szegedy2015-ys,He2016-kt}), finding
the right architecture and associated hyper-parameters, known as \emph{model
search} has kept humans ``in-the-loop.'' Traditionally, the deep learning
community has resorted to techniques such as
grid search and random search \cite{Bergstra2012-rh}.
In recent years, model-based search, in particular, Bayesian Optimization (BO),
has become the preferred technique in many deep learning applications
\cite{Shahriari_undated-vb}.

It is common to apply BO to the
search over various architectural hyper-parameters (e.g.~number of layers,
number of hidden units per layer) but applying BO to search the space of complete
network architectures is much more challenging. Modeling the space of network
architectures as a discrete topological space is equivalent to making random choices between architectures during the
BO procedure - as each architecture variant in the search space would be equally similar. To exploit architectural similarities we require some distance metric between architectures to be quantified.

To this end, we introduce a flexible family of
\emph{graph-induced kernels} which can effectively quantify the
similarity or dissimilarity between different network architectures. We are the first to explore
hierarchical structure learning using BO with a focus on multimodal fusion DNN
architectures. We demonstrate empirically on the Cornell Human Activity \cite{Sung11humanactivity}
Recognition Dataset, and the Montalbano Gesture Recognition Dataset \cite{Escalera_ECCV2014} that the
optimized fusion structure found  using our approach is on-par with
an engineered fusion structure.

\section{Graph-Induced Kernels for Bayesian Optimization}\label{sec:GraphInducedKernels}
Let $S$ be a discrete space. Suppose we want to find $\max\{f(x)|x \in S\}$ where $f$ is some non-negative real-valued
objective function that is expensive to evaluate, such as the classification accuracy on a validation set of a trained
net $x \in S$. We will find an optimal $x^* \in S$ using Gaussian Process-based Bayesian Optimization. More formally,
at each point during the optimization procedure, we have a collection of known pairs $(x^{(i)}, y^{(i)})$ for $i = 1,
\dots, m$, where $y^{(i)} = f(x^{(i)})$. We want to use Gaussian Process regression to model this data, and then use
that to choose the next $x \in S$ to evaluate. To fit a Gaussian Process we need to define a kernel function on the
discrete domain $S$.

\noindent{\bf Radial Kernels:} Let $k$ be a kernel on $S$. We say that $k$ is \emph{radial} when there exists a metric
$d$ on $S$ and some real shape function $r$ such that $k(x, y) = r(d(x, y))$. The kernel could also be described as
\emph{radial with respect to the metric $d$}. For example, the Gaussian and exponential kernels on $ \mathbb{R}^n$ are
both radial with respect to the standard Euclidean metric.

\noindent{\bf Graph-Induced Kernels:} Let $G = (S, E)$ be an undirected graph,
let $d$ be its geodesic distance metric\footnote{The \emph{geodesic distance}
between two vertices in a graph is the number of edges in a shortest path
connecting them.},
and let $r$ be some real shape function. We then define \emph{the kernel $k$,
induced by graph $G$, and shape $r$},
to be $k(x, y) = r(d(x, y))$. For example, choosing the Gaussian shape function gives $k(x, y) = \exp\{-\lambda \cdot
[d(x, y)]^2\}$, where $x, y \in S$ and $\lambda > 0$ is a parameter of the kernel. If the graph edges are assigned
costs, those costs can be treated as the parameters of the kernel instead of $\lambda$.

To apply BO to $\max\{f(x)|x \in S\}$ where $S$ is discrete, we design a graph $G$ that respects the topology of the
domain $S$ and choose a shape function $r$, inducing a kernel on $S$ that we can use to fit a Gaussian Process to the
collection of known pairs $(x^{(i)}, y^{(i)})$ for $i = 1, \dots, m$, where $y^{(i)} = f(x^{(i)})$. This approach is
desirable because it reduces the task of defining a kernel on $S$ to the much simpler task of designing the graph $G$.
This enables the user to flexibly design kernels that are customized to the diverse domain topologies encountered
across a variety of applications.

For example, consider the problem of choosing the best deep multimodal fusion architecture for classification. In this
case, each element of the domain $S$ might be a tree data structure describing a neural network architecture, with the
graph $G$ describing neighbor relationships between similar architectures. For a particular architecture $u \in S$,
each possible modification of the architecture yields a neighboring architecture $v \in S$, where $\{u, v\}$ is an edge
in $G$. To accommodate different modifications to the network structure, each
modification type $t$ can have a corresponding edge weight parameter $w_t$,
such that the graph-induced kernel could be parameterized to respect the
different types of modifications.


\noindent{\bf Network Architecture:} The deep neural network that was used in this work was adapted from the
tree-structured architecture reported in \cite{Neverova_PAMI}. The tree structured network architecture has a
multi-stage training procedure. In the first stage, separate networks are trained to learn modality-specific
representations. The second stage of training consists of learning a multimodal shared representation by fusing the
modality-specific representation layers. We used identical structure and hyper-parameters as reported in that paper for
each modality-specific representation learning layers that are typically pre-trained until convergence.

\begin{figure}[htbp]  \begin{subfigure}[t]{0.4\textwidth}
		\centering \includegraphics[scale=0.5]{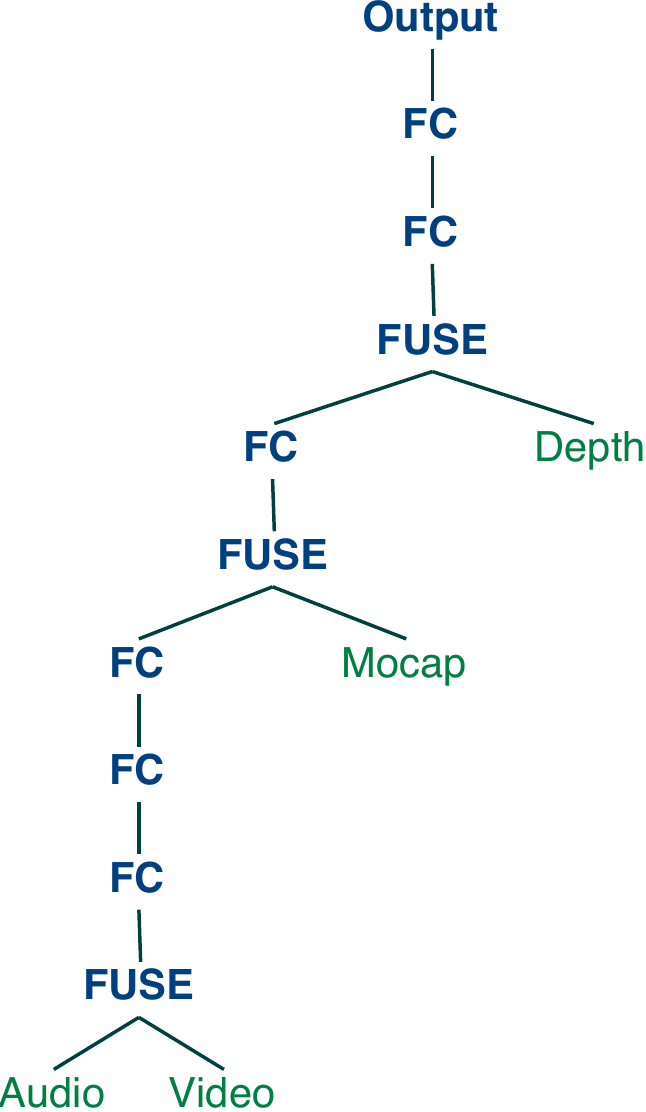}
		\caption{Fusion arch. variant 1}
        \label{fig:archA2}
		\end{subfigure}
	\hfill 
	\begin{subfigure}[t]{0.4\textwidth}
		 \centering\includegraphics[scale=0.5]{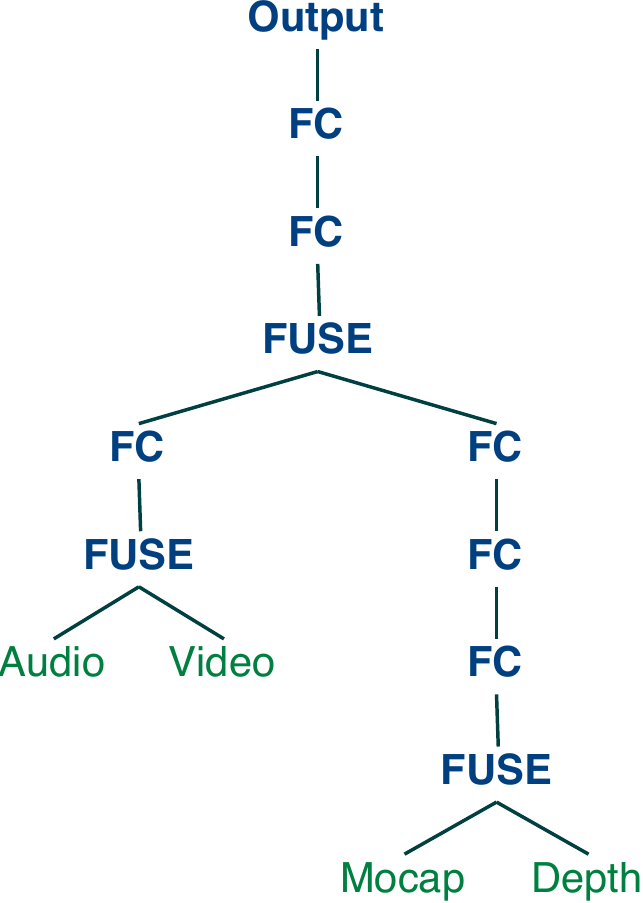}
		\caption{Fusion arch. variant 2}
		\label{fig:archB2} \end{subfigure} \caption{Two
			variants of
		multimodal fusion architectures} \end{figure}

We generalize the fusion strategy to consider $n$-ary fusions between any
number of modality-specific or merged-modality network pathways.
The search space is constructed by adding fully-connected (FC) layers after any
input node or fusion nodes.
Figs.~\ref{fig:archA2} and~\ref{fig:archB2} depict
two possible multimodal fusion architectures with different fusion depths and
orders of fusion.

\noindent{\bf Graph Design:} To apply BO to the problem of finding the best
multimodal fusion architecture (a \emph{net} for brevity), we design a graph
$G$ where the nodes are nets and then use the kernel induced by $G$ and a
shape function $r$.  To design $G$, we first formalize the domain $S$ of nets, then we define the edges of $G$. We
encode a net as a pair $(T, D) \! \in \! S$ where $T$ is a nested set describing the order in which modalities are
fused and $D$ is a map from subtrees (nested sets) to the number of subsequent FC layers.

We define two nets $(T, D), (T', D') \in S$ to be neighbors in $G$ if and only
if exactly one of the following statements holds:
\begin{enumerate}
\item $T'$ can be constructed from $T$ by either adding or removing a single
fusion (while keeping the same set of modalities);
\item $T'$ can be constructed from $T$ also by changing the position of one of
the modalities in the fusion hierarchy, by shifting its merging point to either
earlier or later fusion;
\item $D'$ can be constructed from
$D$ by incrementing or decrementing its total number of FC layers.
\end{enumerate}
Pairing the Gaussian shape function with this completed
definition of $G$ induces the kernel we use during BO to find the optimal
architecture $x \in S$, which can be parameterized by setting the weights of
those edges to be $w_T, w_D > 0$. In our experiments, we simply set $w\_T, w\_D
= 1$
\section{Results and Discussion}

We validated the efficacy of our approach on two datasets:

\noindent{\bf Cornell Human Activity (CAD-60)
Dataset \cite{Sung11humanactivity}:} consists of 5 descriptor-based modalities
derived
from RGB-D video and the objective is to classify over 12 human activity
classes. For each net that was evaluated, we computed the average test accuracy
across 4 cross-validation dataset subsets, yielding a generalized measure of
accuracy for a given net.

\noindent{\bf Montalbano Gesture Recognition Dataset \cite{Escalera_ECCV2014}
:}
is a much larger
dataset compared to CAD-60. It consists of 4 modalities: RGB video,
depth video, mocap, and audio. The objective is to classify and localize over
20 communicative gesture categories.

\begin{figure}[t]
	\centering
	\begin{minipage}{\textwidth}
		\begin{subfigure}[t]{0.5\linewidth}
			\includegraphics[width=\textwidth]{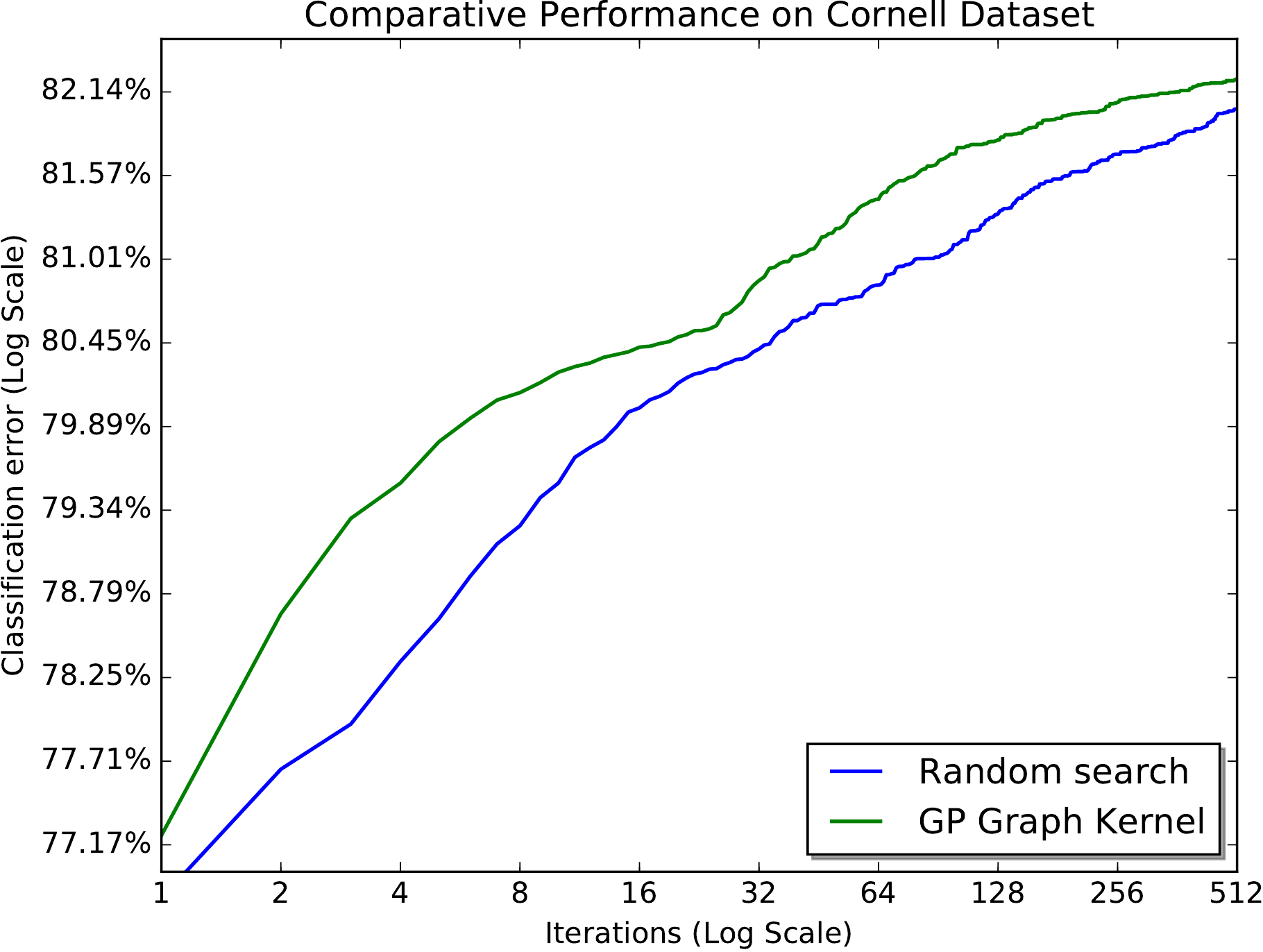}
			\caption{Cornell Experiments}
			\label{fig:cornell-algo-perf-loglog}
		\end{subfigure}
		\hfill
		\begin{subfigure}[t]{0.5\linewidth}
			\includegraphics[width=\textwidth]{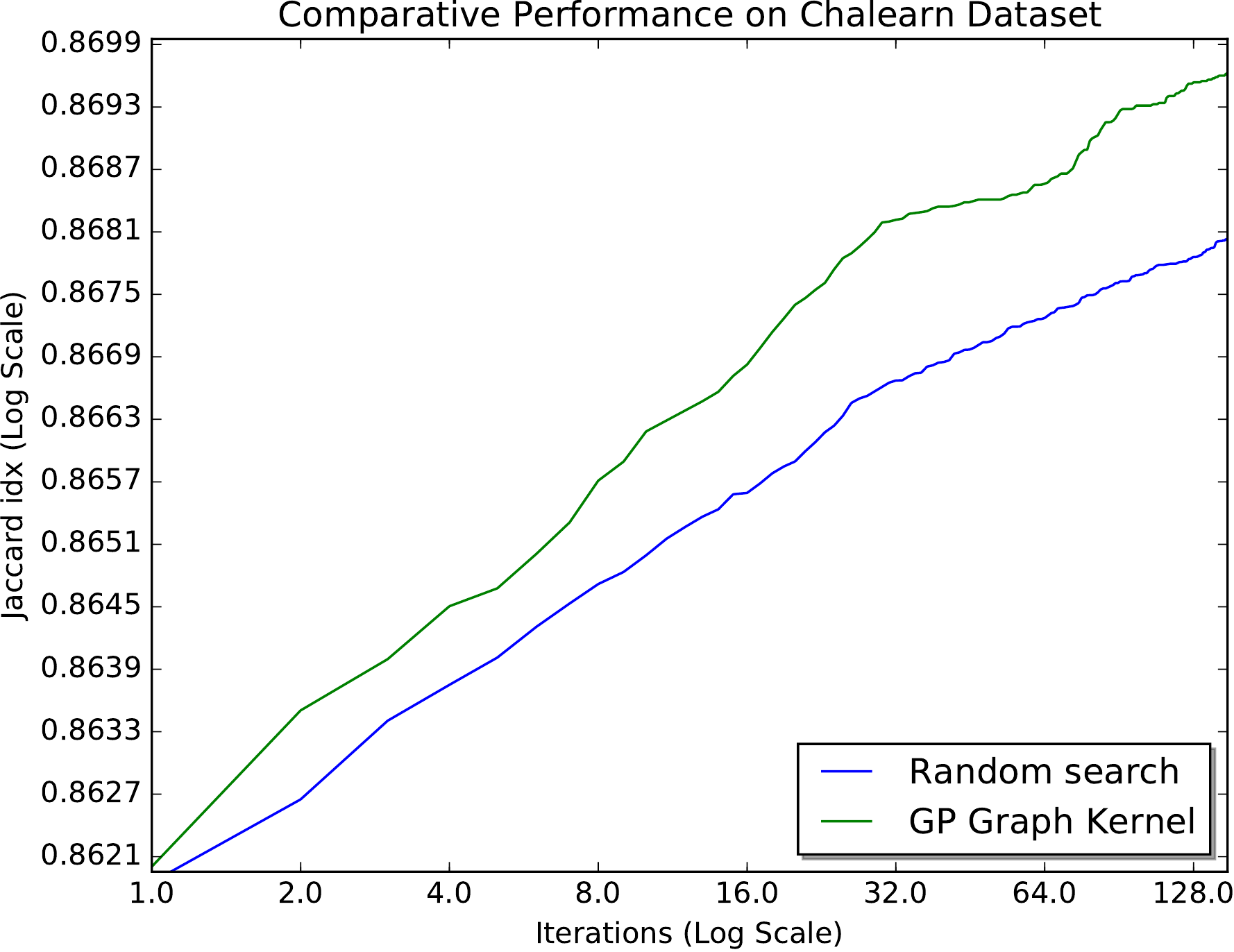}
			\caption{Montalbano Experiments}
			\label{fig:chalearn-algo-perf}  
		\end{subfigure}
		\vspace{-10pt}
		\caption{Comparative performance of the proposed method versus the random search method averaged over 100 runs.}
	\end{minipage}
	\vfill\vspace{5pt}%
	\begin{minipage}{\textwidth}
		\begin{subfigure}[t]{0.5\linewidth}
			\includegraphics[width=\textwidth]{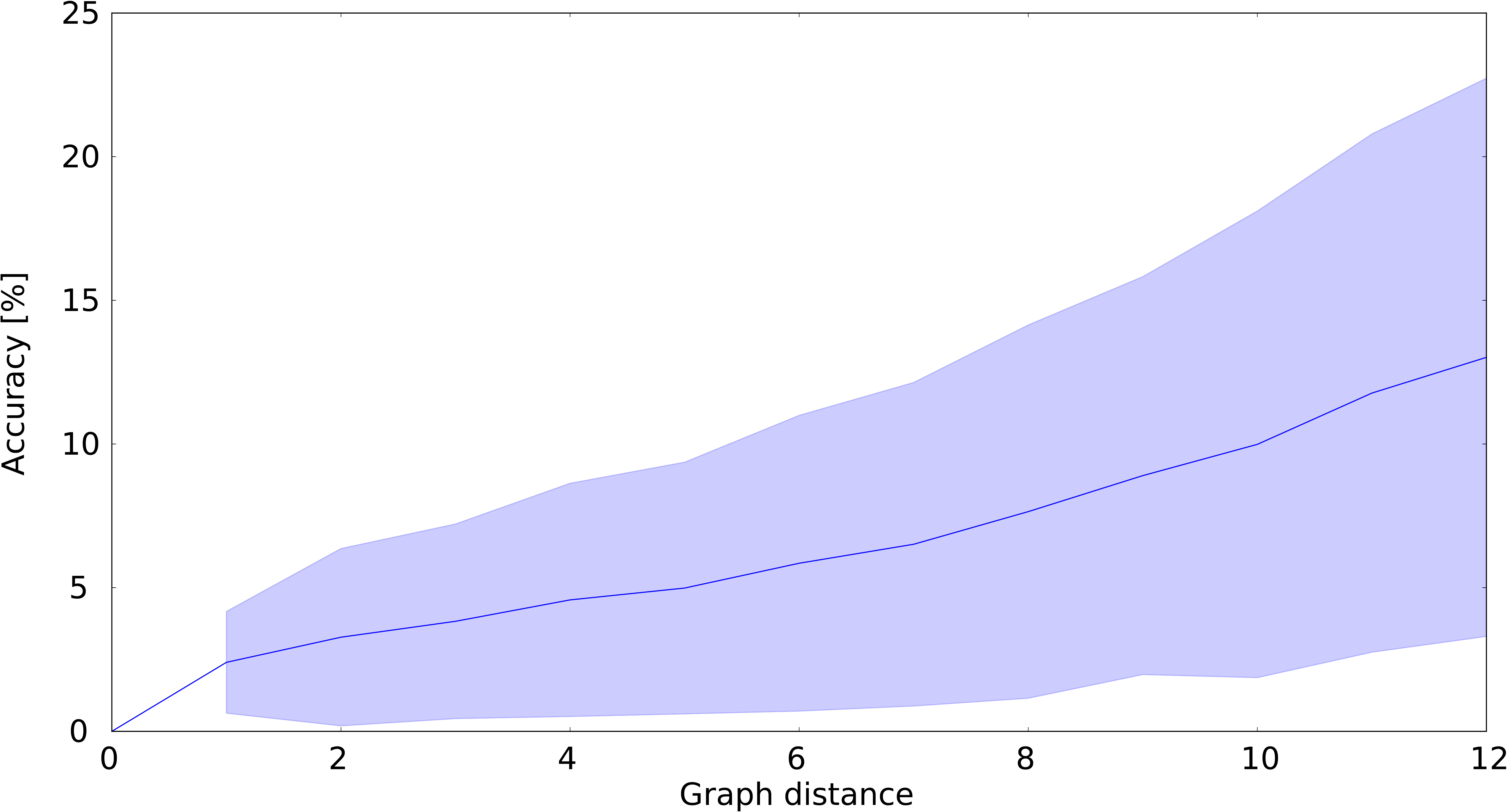}
			\caption{Cornell Experiments}
			\label{fig:cornel-graph-dist-std}
		\end{subfigure}\hfill 
		\begin{subfigure}[t]{0.5\linewidth}
			\includegraphics[width=\textwidth]{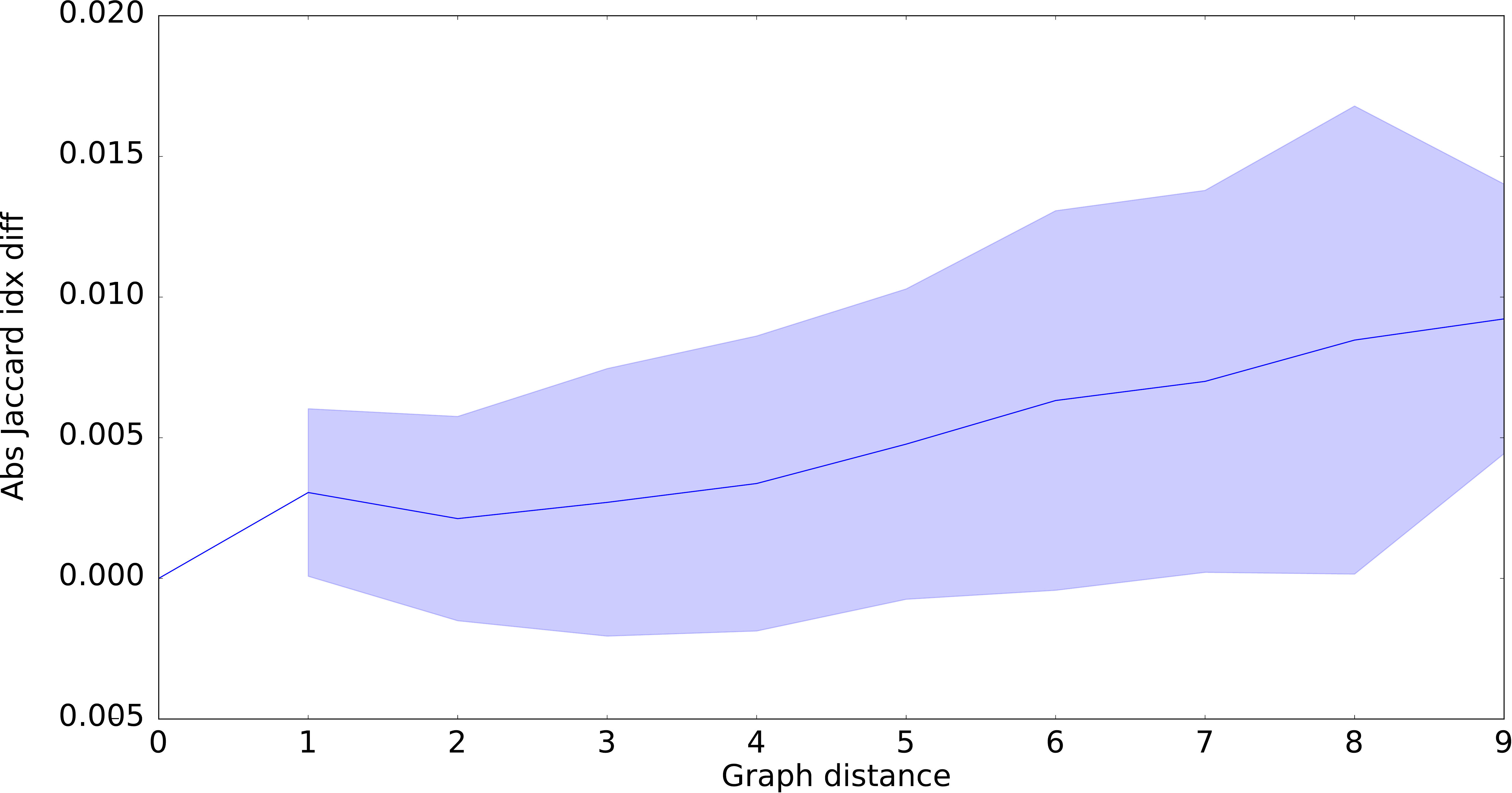}
			\caption{Montalbano Experiments}
			\label{fig:chalearn_graph_jac_mean_std}
		\end{subfigure}
	\end{minipage}
	\vspace{-10pt}
	\caption{The average absolute test accuracy difference as a function of graph kernel distances.}
	\vspace{-10pt}
\end{figure}

We integrated our Graph-induced kernel with a GP--based BO framework
\cite{gpy2014} and compared it with random search for  fusion structure
optimization. The multimodal network architecture was implemented in Lasagne
\cite{lasagne}. We assumed sample noise with a variance of 1.0 for our
normalized inputs. The random search \cite{Bergstra2012-rh} is the baseline that
has been shown to be in line with human performance for the same number of
trials. Fig.~\ref{fig:cornell-algo-perf-loglog} shows the performance of those
two methods averaged over 100 runs. Our method can find an architecture with the
same classification error in 2$\times$ less iterations than the random search.
For example, to find an architecture that achieves 19.7\% validation error, our
approach only needed around 8 iterations, while random search required 18
iterations. Fig.~\ref{fig:cornel-graph-dist-std} shows the average absolute test
accuracy difference obtained as a function of the respective graph kernel
distances computed. The strictly positive trend of this plot suggests that the
metric incorporated into our graph kernel captures enough information about the
search space to correctly evaluate the real distance between network structures.
Fig.~\ref{fig:chalearn-algo-perf} shows the number of iterations needed to find
a network structure that produces good test performance for the Montalbano
dataset. Our proposed technique achieved up to $5\times$ speedup compared to
random search. Fig.~\ref{fig:chalearn_graph_jac_mean_std} shows a similar
positive trend to that seen in CAD-60. Despite having a tight variance between 
performances of different architectures for this dataset, our graph-induced 
kernel provided sufficiently high signal to noise ratio to be usable for 
structure optimization. 

\section{Conclusion and Future Work}\label{sec:conclusion}
In this work, we have proposed a novel graph-induced kernel approach in which
easily-designed graphs can define a kernel specialized for any discrete domain.
To demonstrate its utility,  we have cast a deep multimodal fusion
architecture search as a discrete hyper-parameter optimization problem. We
demonstrate that our method could optimize the network architecture leading to
accuracies that are at par or slightly exceed those of manually-designed
architectures \cite{Neverova_PAMI}
while evaluating between 2-5$\times$ less architectures than random search on 2 challenging
human activity recognition problems.

In our future work, we will extend this approach to non GP-based
alternatives to BO optimization techniques such as TPE\cite{Bergstra2011-fz}
and  SMAC\cite{hutter2011sequential}, in addition to
implementing different graph kernels and examining their relative
performance.

\section*{Acknowledgments} This research is partially funded by the Defense
Advanced Research Projects Agency (DARPA) and the Air Force Laboratory (AFRL).
We would also like to thank the NSERC and Canada Foundation for Innovation for
infrastructure funding as well as NVIDIA Corp.~for contributing an NVIDIA Titan
X GPU.

%
\end{document}